\documentclass[11pt]{article}

\usepackage[final]{acl}

\usepackage{times}
\usepackage{latexsym}

\usepackage[T1]{fontenc}

\usepackage[utf8]{inputenc}

\usepackage{microtype}

\usepackage{inconsolata}

\usepackage{graphicx}

\title{LipoAgent: Coordinating Fine-Tuned LLM Agents for Safer Lipid Design}

\author{Leshu Li$^{1}$\thanks{Equal contribution.},
An Lu$^{2}$\footnotemark[1], Haiyu Wang$^{1}$\thanks{Equal contribution.}, Zhibin Feng$^{1}$\footnotemark[2], Conghui Duan$^{1}$\footnotemark[2], Qing Bao$^{2}$, \\ 
\textbf{Zongmin Zhao$^{2}$, Sai Qian Zhang$^{1}$
} \\
\text{$^{1}$New York University, USA, $^{2}$University of Illinois Chicago, USA} \\
\texttt{\{ll5914,sai.zhang\}@nyu.edu}, \texttt{\{luan1994,zhaozm\}@uic.edu}\\
}

\usepackage{array}
\usepackage{xcolor}
\usepackage{amssymb}
\usepackage{pifont} 
\usepackage{amsmath}
\usepackage{amssymb}
\usepackage{amsfonts}
\usepackage{booktabs}        
\usepackage{graphicx}        
\usepackage[table]{xcolor}   
\usepackage{pifont}          

\newcolumntype{M}[1]{>{\centering\arraybackslash}m{#1}}

\definecolor{darkgreen}{RGB}{0,100,0}

\newcommand{\cmark}{{\color{darkgreen}\bfseries\ding{51}}} 
\newcommand{\xmark}{{\color{red}\bfseries\ding{55}}}   

\newcommand{\agentname}{LipoAgent}
\newcommand{\dataname}{TransLipid}

\begin{document}
\pagestyle{empty}
\maketitle
\begin{abstract}
Lipid nanoparticles (LNPs) are among the most clinically mature platforms for nucleic acid delivery, yet designing lipids that are both effective and biologically safe remains a major bottleneck. In practical screening, toxicity is a decision-level constraint: if a lipid is toxic, its efficiency prediction is clinically irrelevant.
We propose \agentname{}, a safety-aware multi-agent LLM framework for lipid discovery. \agentname{} combines domain-specific fine-tuning with a conditional prediction objective that enforces toxicity as a prerequisite for efficiency prediction, and further improves reliability via multi-agent verification with lightweight human oversight when disagreement persists.
Across multiple foundation models, \agentname{} achieves an average 32\% relative improvement in mRNA transfection efficiency prediction compared with other reported models for lipid design.
~\textbf{Wet-lab validation} confirms that virtual screening rankings reliably translate to biological transfection outcomes. The code is publicly available at \url{https://github.com/SAI-Lab-NYU/LipoAgent.git}.
\end{abstract}

\section{Introduction}

\begin{figure*}[t]
  \centering
  \includegraphics[width=0.9\textwidth]{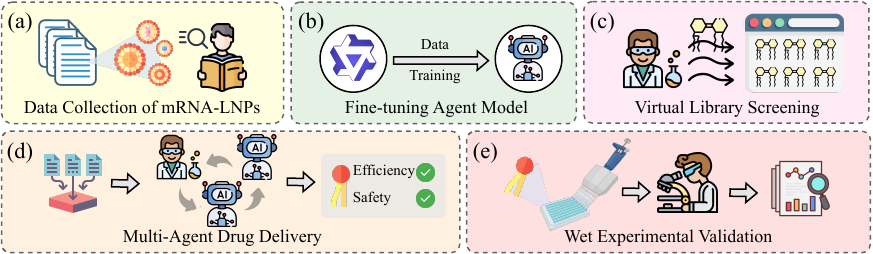}
  \vspace{-0.5em}
  \caption{Overview of \agentname{}, with a safety-aware multi-agent LLM framework for lipid discovery.}
  \vspace{-1em}
  \label{fig:Intro_compare}
\end{figure*}

Messenger ribonucleic acid (mRNA) therapeutics have attracted increasing attention due to their ability to regulate disease-related gene expression~\cite{ann_intro1,An_1}.
The remarkable success of coronavirus disease 2019 (COVID-19) mRNA vaccines, such as BNT162b2 and mRNA-1273, has highlighted the clinical potential of mRNA-based therapies~\cite{an_intro2}.
Lipid nanoparticles (LNPs) play a pivotal role in enabling effective mRNA delivery, owing to their favorable biocompatibility, high encapsulation efficiency, and scalable manufacturing processes~\cite{an_intro3,ann_add,An_2}.
As the key functional components of LNPs, ionizable lipids critically determine mRNA transfection efficiency, which is strongly impacted by their molecular structures~\cite{An_3,An_4}.

Despite these advances, identifying new lipid candidates that achieve both high delivery efficiency and acceptable safety remains a major bottleneck~\cite{An_5}.
Traditionally, optimal lipids are discovered through large-scale combinatorial synthesis followed by extensive in vitro and in vivo screening.
Such workflows are extremely time-consuming and resource-intensive, often requiring months of experimental effort and biological validation~\cite{an_intro5}.
As a result, only a small fraction of the vast chemical design space can be practically explored, limiting both discovery speed and diversity.

Recent progress in data-driven modeling has opened new opportunities to accelerate lipid discovery.
In particular, large language models (LLMs) have demonstrated strong reasoning and knowledge integration capabilities, enabling promising results in molecular generation and property prediction across the drug discovery pipeline~\cite{survey1,survey2,ReACT,Researchagent,DrugAgent,chemcrow}.
These successes suggest that LLMs could serve as powerful assistants for navigating complex chemical spaces and prioritizing candidate molecules~\cite{An_6}.
However, directly applying general-purpose LLMs to lipid design presents fundamental challenges.

First, without domain-specific fine-tuning, LLMs often exhibit limited predictive accuracy for biochemical properties such as mRNA transfection efficiency.
More critically, existing LLM-based approaches typically lack explicit safety modeling.
Toxicity is frequently treated as a post hoc filtering step rather than an integral part of the decision process.
This design choice can lead to unsafe high-confidence predictions, where a lipid is recommended as highly efficient despite exhibiting unacceptable toxicity, thereby introducing substantial experimental risk and wasted validation effort~\cite{safetysurvey}.


To realize this paradigm, we propose \textbf{\agentname{}}, a safety-aware multi-agent LLM framework for lipid discovery.
As illustrated in Figure~\ref{fig:Intro_compare}, \agentname{} integrates domain-specific fine-tuning, conditional prediction, and multi-agent verification into a unified pipeline.
The framework consists of two coordinated agents.
The \textit{Predictor agent} jointly predicts molecular toxicity and mRNA transfection efficiency while producing structured reasoning traces and confidence estimates.
The \textit{Verifier agent} examines low-confidence predictions by checking the consistency between predicted scores and their corresponding explanations.
Crucially, we introduce a conditional loss mechanism during both training and inference: when a molecule is predicted to be toxic, the model directly outputs an ``unsafe'' decision and halts efficiency prediction.
This design enforces toxicity as a prerequisite for efficiency modeling and prevents the system from recommending ``efficient but toxic'' lipid candidates.
To support systematic and reproducible evaluation, we further construct and release a new dataset, \textbf{\dataname{}}.
The dataset is manually curated and normalized from multiple published studies and comprises structure--efficiency--toxicity triplets, providing a unified base for training and evaluating safety-aware molecular models.
In summary, our contributions are as follows:
\newcolumntype{L}[1]{>{\raggedright\arraybackslash}m{#1}} 
\newcolumntype{C}[1]{>{\centering\arraybackslash}m{#1}}   

\begin{table}[t!]
\centering
\small
\renewcommand{\arraystretch}{1}
\begin{tabular}{L{1.4cm} C{0.8cm} C{0.8cm} C{1.3cm} C{1.3cm}}
\hline
\multicolumn{1}{C{1.4cm}}{\textbf{Method}} &
\textbf{Multi-Agent} & \textbf{Fine-tuning} & \textbf{Human Feedback} & \textbf{Toxicity Detection} \\
\hline
ReAct            & \xmark & \xmark & \xmark & \xmark \\
ResearchAgent    & \cmark & \xmark & \xmark & \xmark \\
ChemCrow         & \cmark & \xmark & \cmark & \xmark \\
DrugAgent        & \cmark & \xmark & \cmark & \xmark \\
\textbf{LipoAgent} & \cmark & \cmark & \cmark & \cmark \\
\hline
\end{tabular}
\caption{Comparison between LipoAgent and existing LLM-based systems for molecular and drug discovery. Dark-green bold checkmarks indicate supported capabilities; red bold crosses indicate missing components.}
\label{tab:comparison}
\end{table}

\noindent\textbf{Safety-first multi-agent LLM framework}: We propose \agentname{}, a multi-agent system for toxicity-aware mRNA transfection efficiency prediction in lipid discovery.

\noindent\textbf{Conditional loss and decision-level safety modeling:} We design a conditional loss function that enforces toxicity as a prerequisite for efficiency prediction during both training and inference, reducing unsafe false-positive recommendations.

\noindent \textbf{\dataname{} dataset:} We construct and release \dataname{}, a curated dataset that integrates lipid molecular structures with experimentally reported transfection efficiency data and toxicity annotations when available, enabling systematic evaluation of safety-aware molecular reasoning.

\noindent\textbf{End-to-end experimental validation:} Extensive quantitative evaluations show that \agentname{} performs competitively against other reported models for lipid design. Real-cell transfection and cytotoxicity experiments further confirm the accuracy of \agentname{}'s predictions and their relevance to practical biological settings.

\section{Background and Related Works}

\subsection{Lipid Design}
The structural features of lipid materials play a decisive role in mRNA delivery performance~\cite{lnp_mrna1,lnp_mrna2}.
Subtle variations in molecular architecture can markedly influence particle formation, cellular uptake, endosomal escape, and ultimately protein expression~\cite{bg_1}.
Lipid-library screening is a commonly used strategy for identifying effective materials for mRNA delivery.
However, experimental throughput is fundamentally limited by cost, labor, and resource demands, which severely constrain the scale of combinatorial exploration~\cite{LNP_AI1,LNP_AI2}.
As a result, promising lipid candidates may remain undiscovered, and the overall optimization process is often slow and inefficient.

These limitations motivate computational approaches that can design lipid candidates in silico.
Importantly, such approaches must simultaneously account for delivery efficiency and biological safety, as high transfection efficiency alone is insufficient for clinical applicability~\cite{AI_Safety1, AI_Safety2}.

\subsection{LLMs for Drug Discovery and Delivery}
In recent years, large language models (LLMs) have been increasingly applied to drug discovery and molecular design owing to their strong reasoning, planning, and knowledge integration capabilities~\cite{AI4S1,AI4S2,AI4S3}.
The early ReAct framework~\cite{ReACT} introduced a reasoning--acting paradigm that enables models to alternate between thought generation and tool invocation for stepwise problem solving.
However, ReAct lacks domain-specific adaptation and does not consider biochemical safety.

Building on this idea, ResearchAgent~\cite{Researchagent} incorporates a multi-agent structure for hypothesis generation, experiment planning, and literature-based reasoning.
Despite its effectiveness in scientific knowledge synthesis, it remains limited to textual reasoning and does not perform quantitative prediction of molecular properties.
ChemCrow~\cite{chemcrow} further advances LLM-based chemistry applications by integrating domain-specific tools such as molecule generators and property calculators, demonstrating the potential of human--AI collaboration.
Nevertheless, it relies heavily on post hoc expert validation and lacks automated toxicity modeling.

More recently, DrugAgent~\cite{DrugAgent} proposed a multi-agent system consisting of a planner and an instructor to automate machine learning programming for drug discovery.
While it demonstrates agent coordination capabilities, it does not incorporate domain-specific fine-tuning and lacks molecular-level safety assessment.

As summarized in Table~\ref{tab:comparison}, existing LLM-based frameworks vary in their use of multi-agent coordination and human feedback.
However, most approaches treat toxicity as a post hoc filtering step rather than a decision-level constraint.
As a result, efficiency prediction and reasoning are not conditioned on safety, which can lead to high-confidence but unsafe molecular recommendations.
In contrast, our proposed framework, \agentname{}, integrates domain-specific fine-tuning, safety-aware prediction, and multi-agent verification within a unified system to jointly optimize lipid transfection efficiency and safety.



\begin{figure*}[t!]
  \includegraphics[width=\textwidth]{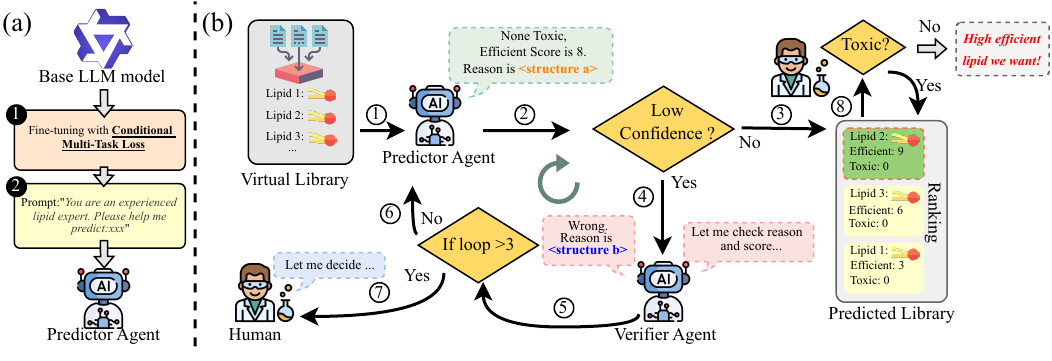}
  \vspace{-1.5em}
    \caption{Overview of the LipoAgent framework. (a) Fine-tuning and prompting pipeline for constructing the predictor agent from a base LLM. (b) Multi-agent collaboration in LipoAgent, where agents coordinate with human feedback to iteratively filter and refine candidates toward high-efficiency lipids.}
\vspace{-1em}
  \label{fig:framework}
\end{figure*}

\subsection{Multi-agent LLM}
Multi-agent LLM frameworks aim to improve reasoning reliability and scalability by decomposing complex tasks into cooperative submodules.
Systems such as CAMEL~\cite{CAMEL}, Voyager~\cite{Voyager}, and Reflexion~\cite{Reflexion} demonstrate how agent interaction, self-reflection, and feedback loops can enhance task performance in scientific reasoning.

Our work adopts a similar multi-agent philosophy but applies it to safety-critical molecular discovery.
In \agentname{}, the Predictor agent focuses on joint toxicity and efficiency prediction, while the Verifier agent performs consistency checking between predicted scores and their corresponding reasoning traces.
This verification mechanism is particularly important in high-confidence regimes, where single-agent reasoning may still produce incorrect but persuasive explanations.
By incorporating safety-aware multi-agent verification, \agentname{} bridges the gap between LLM-based molecular reasoning and experimental reliability.

\section{Methodology}
We propose \agentname{}, a multi-agent framework for safe, reliable, and interpretable LLM-based lipid discovery for mRNA delivery, comprising two agents and a fine-tuned predictor.
(1) a \textit{Predictor Agent} that predicts lipid toxicity and delivery efficiency while generating textual reasoning and uncertainty estimation, and 
(2) a \textit{Verifier Agent} that inspects low-confidence predictions and validates the logical consistency between the predicted score and its explanation.
Through a human-in-the-loop feedback loop, LipoAgent supports iterative correction and continual refinement, as shown in Figure~\ref{fig:framework}.

\subsection{Fine-Tuning Strategy for Predictor Agent}

The Predictor Agent is trained with a \textit{Conditional Multi-Task Loss} that jointly optimizes toxicity classification and efficiency prediction. We apply LoRA~\cite{hu2022lora} to the projection layers of the Predictor Agent while freezing all other parameters, substantially reducing training cost.
Given a mini-batch of lipid inputs $\{\mathbf{x}_i\}_{i=1}^{N}$, the model outputs a toxicity logit $z_i^{\text{tox}}$ and an efficiency logit vector $z_i^{\text{eff}}$ over discrete transfection efficiency levels. The output probabilities are:

\[
\small
p_{\text{tox},i} = \sigma\!\big(z_i^{\text{tox}}\big), 
\qquad
p_{\text{eff},i} = \mathrm{Softmax}\!\big(z_i^{\text{eff}}\big).
\]

\paragraph{Toxicity Loss.}
Toxicity prediction is trained using binary cross-entropy:
\begin{equation}
\small
\begin{aligned}
\mathcal{L}_{\text{tox}}
= - \frac{1}{N} \sum_{i=1}^{N}
\Big[
& y_i^{\text{tox}} \log \sigma\!\big(z_i^{\text{tox}}\big) \\
& + \big(1 - y_i^{\text{tox}}\big)
\log\!\big(1 - \sigma\!\big(z_i^{\text{tox}}\big)\big)
\Big]
\end{aligned}
\end{equation}
where $y_i^{\text{tox}} \in \{0,1\}$ is the ground-truth label.

\paragraph{Efficiency Loss (Non-toxic Samples Only).}
Efficiency supervision is applied only to non-toxic lipids. Let  
$m_i = 1\{ y_i^{\text{tox}} = 0 \}$  
be a binary mask indicating non-toxic samples. CE(.) is the cross-entropy loss, then the efficiency loss is:
\begin{equation}
\small
\mathcal{L}_{\text{eff}}
= \frac{\sum_{i=1}^{N} m_i \, \mathrm{CE}\big(z_i^{\text{eff}}, y_i^{\text{eff}}\big)}
       {\sum_{i=1}^{N} m_i + \epsilon},
\end{equation}
where $y_i^{\text{eff}} \in \{1,\dots,10\}$ is the discrete efficiency score. $\epsilon$ is a small constant introduced for stability.
The final objective can be expressed as:
\begin{equation}
\small
\mathcal{L}_{\text{total}} 
= \mathcal{L}_{\text{tox}} + \alpha \, \mathcal{L}_{\text{eff}}.
\end{equation}

\subsection{Multi-Agent Verification Framework}

A key component of \agentname{} is an entropy-based confidence score that determines whether additional verification is needed. Given the efficiency distribution $p_{\text{eff}}$, the entropy is computed as:
\begin{equation}
\small
\mathcal{H}(p_{\text{eff}}) 
= - \sum_{k=1}^{10} p_{\text{eff}}^{(k)} \log p_{\text{eff}}^{(k)}.
\end{equation}
We normalize it into a confidence score:
\begin{equation}
\small
\mathrm{Conf}(x)
= 1 - \frac{\mathcal{H}(p_{\text{eff}})}{\log 10},
\end{equation}
where $\mathcal{H}(p_{\text{eff}}) \in [0, \log 10]$ for a 10-class distribution, and the division by $\log 10$ normalizes the confidence score to the range $[0,1]$, with higher entropy indicating lower confidence.

\agentname{} forms an iterative verification loop between the Predictor and Verifier Agents. 
The Predictor outputs $(y_{\text{tox}}, y_{\text{eff}}, r_{\text{pred}}, \mathrm{Conf})$, and predictions with $\mathrm{Conf} > \tau$ are accepted directly.  
Samples with $\mathrm{Conf} \le \tau$ are passed to the Verifier for evaluating reasoning--score consistency:
\begin{equation}
\small
y_{\text{ver}} = f_{\text{ver}}(r_{\text{pred}}, y_{\text{eff}}) \in \{0,1\}.
\end{equation}

When inconsistency is detected, structured corrective feedback $r_{\text{corr}}$ is generated and fed back to the Predictor for the next inference round.

\begin{figure}[h]
  \includegraphics[width=1.0\columnwidth]{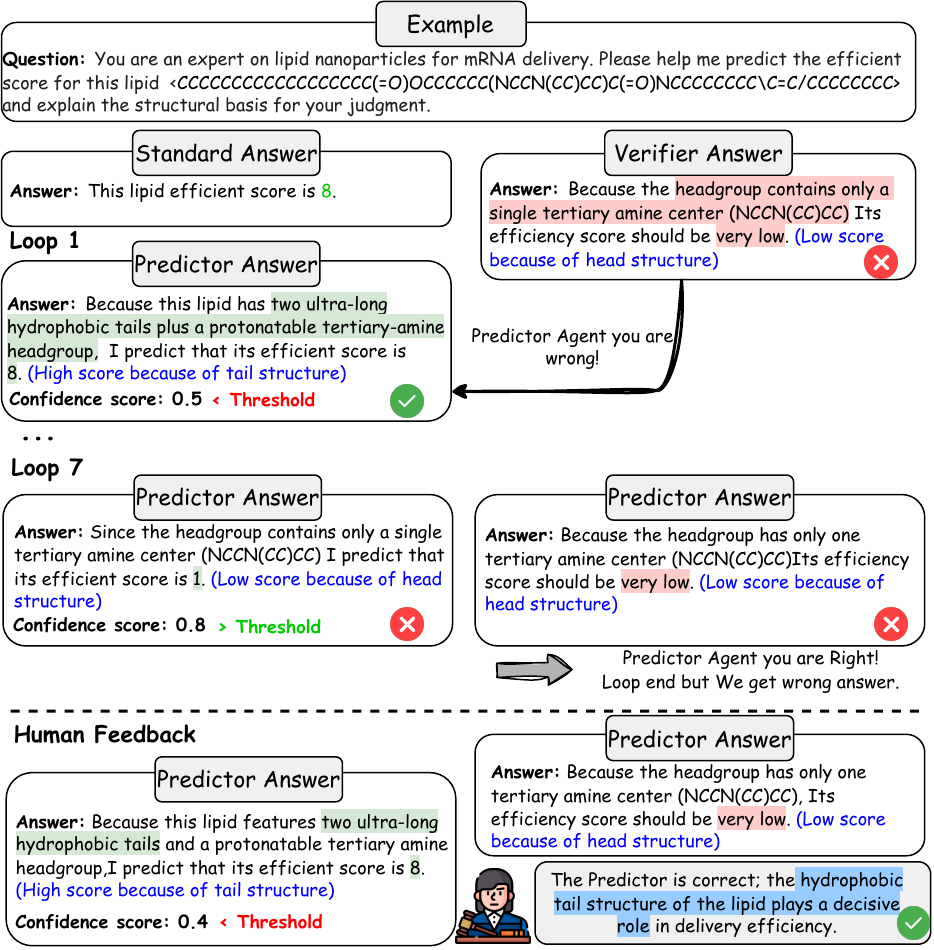}
  \vspace{-1.5em}
    \caption{Multi-agent disagreement in iterative molecular efficiency prediction. During multi-step inference, the two agents may generate inconsistent efficiency estimates, leading to erroneous final selections. Incorporating lightweight human feedback resolves these inconsistencies and improves overall prediction accuracy.}
\vspace{-1em}
  \label{fig:humanfeedback}
\end{figure}
\subsection{Human-in-the-Loop}
To improve safety and reliability, \agentname{} incorporates a human-in-the-loop mechanism. As illustrated in Figure~\ref{fig:humanfeedback}, iterative multi-agent inference can result in disagreement between the Predictor and the Verifier. In some cases, the agents enter repetitive loops without reaching consensus; although a decision may eventually emerge after additional iterations (e.g., more than three rounds), such late-converged predictions are often unreliable or incorrect.
To mitigate this failure mode, we introduce lightweight human feedback. When the two agents fail to resolve their disagreement with sufficient confidence, a human expert provides a final judgment. This intervention prevents error accumulation from prolonged inference loops and improves overall system efficiency and accuracy.



\section{Evaluation}
\begin{table*}[t!]\centering
\caption{Comparison of methods and LLM baselines across efficiency/toxicity metrics.}
\vspace{-0.5em}
\label{tab:methods_comparison}
\resizebox{\linewidth}{!}{
\begin{tabular}{lcccccccc}
\toprule
\hline
\textbf{Methods} & \textbf{Architecture} & \textbf{Fine-tuned} & \textbf{Multi-agent} &
\textbf{Efficiency accuracy} & \textbf{Extreme accuracy$^{a}$} & \textbf{Middle accuracy$^{b}$} &
\textbf{MAE$^{c}$} & \textbf{Toxic accuracy} \\
\hline
AGILE~\cite{AGLE}     & GNN & \xmark & \xmark & 30.80\% & 87.83\%    & 20.10\% & 2.28 & - \\
SCENT~\cite{SCE}    & GNN & \xmark & \xmark & 32.56\%      & 40.33\%      & 28.45\%      & 2.21   & -   \\
LANTERN~\cite{LANTERN}   & MLP & \xmark & \xmark & 42.33\% & 37.22\% & 51.15\% & 2.10 & 58.76\% \\
KnowMol~\cite{KnowMol}   & LLM & \cmark & \xmark & 62.34\%      & 65.67\%      & 63.26\%      & 1.81   & 62.12\%  \\

DrugPilot~\cite{DrugPilot} & LLM & \cmark & \xmark & 53.47\%    & 55.87\%      & 50.86\% & 1.97   & 66.05\%  \\
DrugAgent~\cite{DrugAgent} & LLM & \xmark & \cmark & 61.21\%    & 63.42\%      & 58.80\% & 1.98   & 65.05\%  \\
\hline
Qwen 3-8B~\cite{qwen}                 & LLM & \xmark & \xmark & 60.23\% & 72.34\% & 55.27\% & 1.87 & 70.40\% \\
Qwen 3-8B w/ Fine-tuned      & LLM & \cmark & \xmark & 70.40\% & 86.36\%    & 65.30\% & 1.24 & 90.30\% \\
\cellcolor[HTML]{DAE8FC}\textbf{Qwen 3-8B w/ LipoAgent}
& \cellcolor[HTML]{DAE8FC}LLM
& \cellcolor[HTML]{DAE8FC}\cmark
& \cellcolor[HTML]{DAE8FC}\cmark
& \cellcolor[HTML]{DAE8FC}\textbf{76.80\%}
& \cellcolor[HTML]{DAE8FC}\textbf{89.60\%}
& \cellcolor[HTML]{DAE8FC}\textbf{69.30\%}
& \cellcolor[HTML]{DAE8FC}\textbf{1.09}
& \cellcolor[HTML]{DAE8FC}\textbf{100.00\%} \\
\hline
Qwen 3-32B~\cite{qwen}                & LLM & \xmark & \xmark & 62.56\% & 67.23\% & 72.98\% & 1.85 & 70.40\% \\
Qwen 3-32B w/ Fine-tuned     & LLM & \cmark & \xmark & 86.70\% & 92.31\%    & 75.00\% & 1.21 & 93.30\% \\
\cellcolor[HTML]{DAE8FC}\textbf{Qwen 3-32B w/ LipoAgent}
& \cellcolor[HTML]{DAE8FC}LLM
& \cellcolor[HTML]{DAE8FC}\cmark
& \cellcolor[HTML]{DAE8FC}\cmark
& \cellcolor[HTML]{DAE8FC}\textbf{89.20\%}
& \cellcolor[HTML]{DAE8FC}\textbf{92.87\%}
& \cellcolor[HTML]{DAE8FC}\textbf{84.32\%}
& \cellcolor[HTML]{DAE8FC}\textbf{1.01}
& \cellcolor[HTML]{DAE8FC}\textbf{100.00\%} \\
\hline
ChemLLM~\cite{ChemLLM}                 & LLM & \xmark & \xmark & 11.10\% & 13.21\%      & 10.74\%      & 2.96   & 56.84\%  \\
ChemLLM w/ Fine-tuned       & LLM & \cmark & \xmark & 65.57\% & 69.33\%      & 63.47\%      &1.87  & 72.11\% \\
\cellcolor[HTML]{DAE8FC}\textbf{ChemLLM w/ LipoAgent}
& \cellcolor[HTML]{DAE8FC}LLM
& \cellcolor[HTML]{DAE8FC}\cmark
& \cellcolor[HTML]{DAE8FC}\cmark
& \cellcolor[HTML]{DAE8FC}\textbf{71.41\%}
& \cellcolor[HTML]{DAE8FC}\textbf{76.97\%}
& \cellcolor[HTML]{DAE8FC}\textbf{69.38\%}
& \cellcolor[HTML]{DAE8FC}\textbf{1.33}
& \cellcolor[HTML]{DAE8FC}\textbf{100.00\%} \\
\hline
Llama 3.1-8B~\cite{Llama}               & LLM & \xmark & \xmark & 35.00\% & 38.26\%      & 32.33\%      & 1.99 & 65.33\% \\
Llama 3.1-8B w/ Fine-tuned   & LLM & \cmark & \xmark & 68.83\% & 70.75\% & 54.29\% & 1.57 & 77.40\% \\
\cellcolor[HTML]{DAE8FC}\textbf{Llama 3.1-8B w/ LipoAgent}
& \cellcolor[HTML]{DAE8FC}LLM
& \cellcolor[HTML]{DAE8FC}\cmark
& \cellcolor[HTML]{DAE8FC}\cmark
& \cellcolor[HTML]{DAE8FC}\textbf{72.34\%}
& \cellcolor[HTML]{DAE8FC}\textbf{74.35\%}
& \cellcolor[HTML]{DAE8FC}\textbf{69.38\%}
& \cellcolor[HTML]{DAE8FC}\textbf{1.32}
& \cellcolor[HTML]{DAE8FC}\textbf{100.00\%} \\
\hline
TxGemma-7B~\cite{Txgemma}             & LLM & \xmark & \xmark & 80.20\% & 55.45\%    & 83.60\% & 1.33 & 85.44\% \\
TxGemma-7B w/ Fine-tuned    & LLM & \cmark & \xmark & 89.50\% & 84.38\% & 91.23\% & 1.13 & 92.87\% \\
\cellcolor[HTML]{DAE8FC}\textbf{TxGemma-7B w/ LipoAgent}
& \cellcolor[HTML]{DAE8FC}LLM
& \cellcolor[HTML]{DAE8FC}\cmark
& \cellcolor[HTML]{DAE8FC}\cmark
& \cellcolor[HTML]{DAE8FC}\textbf{94.23\%}
& \cellcolor[HTML]{DAE8FC}\textbf{90.83\%}
& \cellcolor[HTML]{DAE8FC}\textbf{95.23\%}
& \cellcolor[HTML]{DAE8FC}\textbf{0.85}
& \cellcolor[HTML]{DAE8FC}\textbf{100.00\%} \\
\hline
TxGemma-27B~\cite{Txgemma}            & LLM & \xmark & \xmark & 82.34\% & 80.23\% & 83.12\% & 1.25 & 86.70\% \\
TxGemma-27B w/ Fine-tuned   & LLM & \cmark & \xmark & 91.31\% & 89.47\% & 92.48\% & 1.10 & 94.20\% \\
\cellcolor[HTML]{DAE8FC}\textbf{TxGemma-27B w/ LipoAgent}
& \cellcolor[HTML]{DAE8FC}LLM
& \cellcolor[HTML]{DAE8FC}\cmark
& \cellcolor[HTML]{DAE8FC}\cmark
& \cellcolor[HTML]{DAE8FC}\textbf{94.23\%}
& \cellcolor[HTML]{DAE8FC}\textbf{92.34\%}
& \cellcolor[HTML]{DAE8FC}\textbf{95.23\%}
& \cellcolor[HTML]{DAE8FC}\textbf{0.81}
& \cellcolor[HTML]{DAE8FC}\textbf{100.00\%} \\
\hline
\bottomrule
\end{tabular}
}
\raggedright
\footnotesize
$^{a}$ \textit{Extreme accuracy} measures prediction accuracy on lipids whose ground-truth mRNA transfection efficiency scores are at the extremes (i.e., 1, 2, 9, and 10). These cases correspond to either highly ineffective or highly effective lipids and are particularly informative for biological screening, especially for identifying candidates with very high transfection efficiency (scores 9 and 10). \\
$^{b}$ \textit{Middle accuracy} measures prediction accuracy on lipids with intermediate ground-truth efficiency scores ranging from 3 to 8.  \\
$^{c}$ \textit{MAE} (Mean Absolute Error) quantifies the average absolute deviation between the predicted mRNA transfection efficiency score and the ground-truth score across all lipids. Lower MAE values indicate better prediction accuracy.
\end{table*}

The training and evaluation data used in this study are drawn from previously published literature~\cite{dataset1} and organized into a curated dataset termed \textbf{TransLipid}. Specifically, we manually curate approximately 1{,}200 entries from peer-reviewed studies on lipid-based delivery materials, each containing the lipid chemical structure and its experimentally measured mRNA transfection efficiency.
To support toxicity modeling and evaluation, we additionally incorporate 400 toxic lipid or lipid-like molecules from the publicly available \texttt{toxic\_30\_datasets}~\cite{dataset2}.
After integration, TransLipid contains a total of 1{,}600 molecular entries, which are used for subsequent alignment, training, and evaluation.

Since the data originate from multiple independent studies, reported transfection efficiency scores are not directly comparable across sources.
To address this issue, we rescale and align all 1{,}600 samples in TransLipid using a unified evaluation protocol and a consistent scoring data.
This alignment is performed solely to ensure cross-study comparability rather than to optimize model performance.

For all experiments, we use a fixed split of the aligned TransLipid dataset, with 800 samples for training and 800 samples for evaluation.

\subsection{Experimental Setting}
\label{sec:exp_setting}
All experiments are conducted on a single node with 4$\times$ NVIDIA H800 GPUs.
We report averaged results over three independent runs with different random seeds.

\paragraph{Base LLMs and initialization.}
We consider multiple base LLM checkpoints to instantiate the Predictor Agent, including
Qwen3-8B~\cite{qwen}, Qwen3-32B, ChemLLM~\cite{ChemLLM}, Llama~3.1-8B~\cite{Llama}, TxGemma-7B~\cite{Txgemma}, and TxGemma-27B.
All models are initialized from their official pretrained checkpoints and trained under identical experimental settings.

\paragraph{Fine-tuning details.}
We adopt parameter-efficient fine-tuning with LoRA, applied to the attention projection modules
(\texttt{q\_proj} and \texttt{v\_proj}), while keeping all remaining parameters frozen.
The learning rate is set to $2\times 10^{-4}$.
For the conditional multi-task objective in Eq.~(3), we set the loss weight $\alpha=0.1$.

\paragraph{Multi-agent inference and human feedback.}
During inference, the Predictor and Verifier agents interact in an iterative verification loop.
If the loop count exceeds three and the two agents fail to reach a consistent decision on a lipid's mRNA transfection efficiency, lightweight human feedback is triggered to provide a final judgment and terminate the loop.
Human intervention is thus used as a fail-safe mechanism rather than a default inference path.

\paragraph{Baselines.}
We benchmark against diverse architectures on TransLipid, including
GNN-based methods (AGILE~\cite{AGLE}, SCENT~\cite{SCE}), an MLP-based method (LANTERN~\cite{LANTERN}),
LLM-based approaches (KnowMol~\cite{KnowMol}, DrugPilot~\cite{DrugPilot}),
and a multi-agent system (DrugAgent~\cite{DrugAgent}).
Benchmark baselines are evaluated using their released code and pretrained models when available.
Since DrugAgent does not provide an official implementation, we reproduce it following the original paper under the same evaluation protocol.

\subsection{mRNA Transfection Efficiency and Toxicity}
The overall performance of mRNA transfection efficiency and toxicity prediction is summarized in Table~\ref{tab:methods_comparison}.
Across all base LLM backbones, incorporating the proposed \agentname{} framework leads to substantial improvements in both efficiency and toxicity prediction accuracy.
On average, \agentname{} improves prediction accuracy by approximately 32\%, with all enhanced models achieving over 70\% accuracy on both tasks.
These results consistently outperform baselines based on GNNs, MLPs, and prior LLM-based approaches.

Importantly, performance gains arise from complementary components of the framework.
Domain-specific fine-tuning primarily improves overall prediction accuracy,
while multi-agent verification disproportionately enhances reliability on difficult and extreme cases.
As shown in Table~\ref{tab:methods_comparison}, \agentname{} achieves over 85\% accuracy on extreme efficiency values (scores 1, 2, 9, and 10).
Such extreme cases are particularly critical in practical lipid screening, where highly efficient candidates (scores 9 and 10) are prioritized for costly experimental validation and highly inefficient candidates must be reliably filtered out.

\begin{figure}[h]
  \includegraphics[width=1.0\columnwidth]{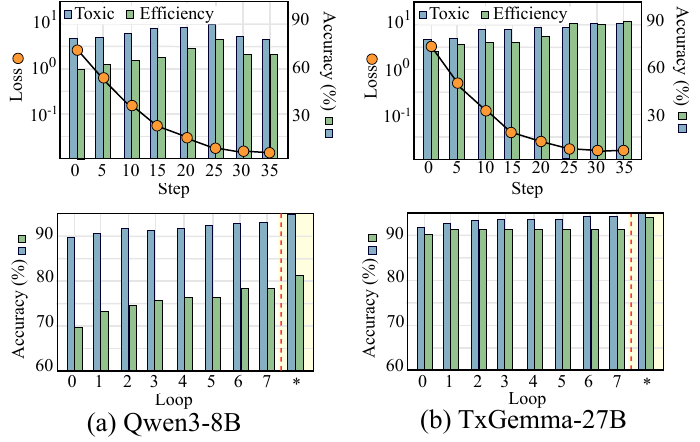}
  \vspace{-1.5em}
  \caption{Training and inference dynamics of \agentname{}.Top: training loss, toxicity accuracy, and mRNA transfection efficiency.Bottom: toxicity and efficiency accuracy across verification loops.“*” indicates the introduction of human feedback.}
  \vspace{-1em}
  \label{fig:exp_finetune}
\end{figure}

Figure~\ref{fig:exp_finetune} further analyzes training and inference dynamics.
During fine-tuning, smaller base models occasionally exhibit initial accuracy gains followed by degradation in later stages, indicating overfitting under limited training data.
Accordingly, for subsequent experiments and wet-lab validation (Section~\ref{sec:wet_validation}), we selected checkpoints with the best validation performance rather than the final training checkpoints.

\begin{figure}[h]
  \includegraphics[width=1.0\columnwidth]{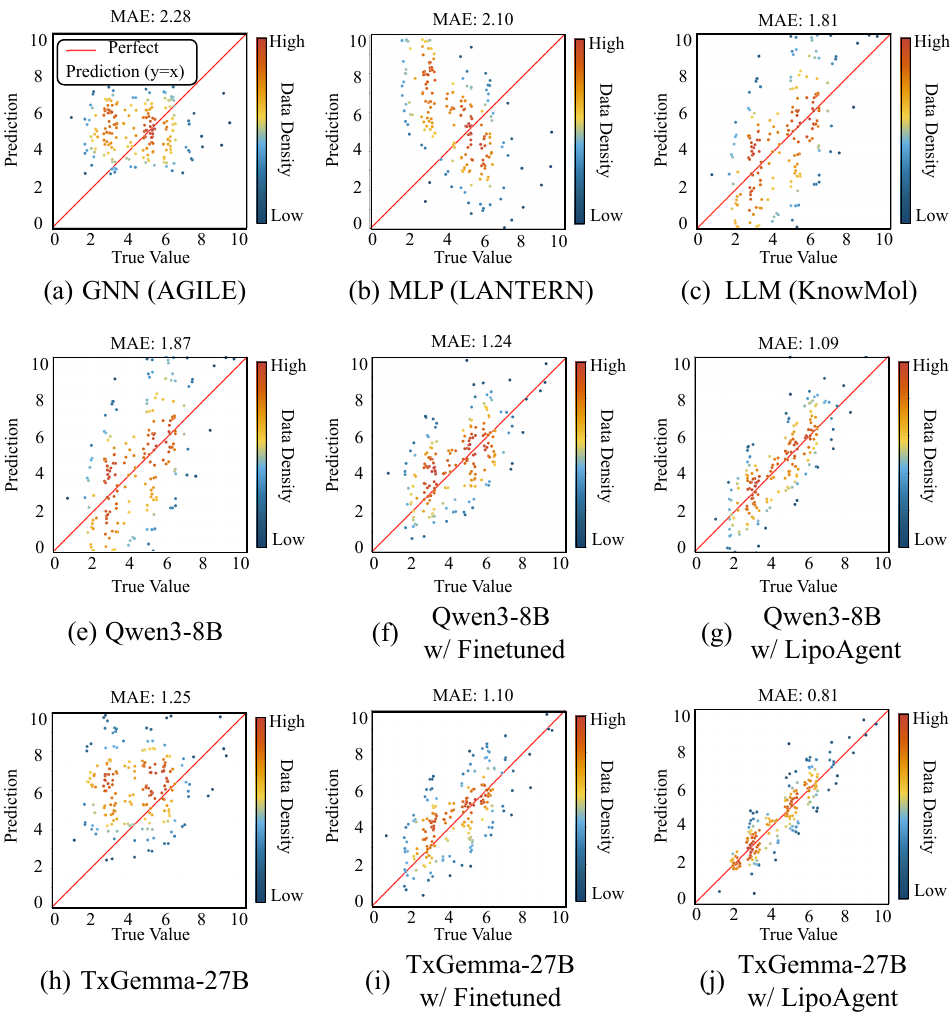}
  \vspace{-1.5em}
  \caption{Prediction distributions of multiple baselines and \agentname{} variants on the test set.
  The red diagonal line ($y=x$) indicates ideal predictions.}
  \vspace{-1em}
  \label{fig:exp_disdribute}
\end{figure}

The figure also highlights the effect of the multi-agent verification loop.
Prediction accuracy for both toxicity and efficiency improves steadily with additional verification rounds and converges after several iterations.
However, toxicity accuracy does not reach 100\% through autonomous multi-agent interaction alone.
In safety-critical drug delivery scenarios, even a small number of false-negative toxicity predictions can pose serious risks.
By introducing lightweight human feedback only when agent disagreement persists, \agentname{} achieves perfect toxicity prediction accuracy while further improving efficiency prediction reliability.

Figure~\ref{fig:exp_disdribute} visualizes prediction distributions across modeling paradigms.
Without domain-specific fine-tuning or verification, most models are biased toward intermediate efficiency scores (typically 4--6), a common failure mode in screening tasks.
With fine-tuning and multi-agent verification, predictions increasingly concentrate along the ideal $y=x$ diagonal.
Notably, \agentname{} improves accuracy at extreme efficiency values, enabling reliable identification of both highly promising and clearly unsuitable lipid candidates.

\subsection{Ablation Study}
\label{sec:ablation}
\begin{table}[t]
\centering
\caption{Ablation study on the timing of human-in-the-loop intervention.}
\label{tab:ablation_loop}
\resizebox{\linewidth}{!}{
\begin{tabular}{c|ccc}
\toprule
\textbf{Loop} & \textbf{Efficiency Acc. (\%)} & \textbf{Toxic Acc. (\%)} & \textbf{MAE} \\
\midrule
No Human &70.40\%  &90.30\%  &1.24  \\
1 &72.56\%  &100.00\%  &1.12  \\
2 &74.20\%  &100.00\%  &1.10  \\
\cellcolor[HTML]{DAE8FC}\textbf{3} &\cellcolor[HTML]{DAE8FC}\textbf{76.80\%}  &\cellcolor[HTML]{DAE8FC}\textbf{100.00\%}  &\cellcolor[HTML]{DAE8FC}\textbf{1.09 } \\
4 &75.32\%  &100.00\%  &1.07  \\
5 &74.01\%  &100.00\% &1.09  \\
6 &74.11\%  &100.00\%  &1.09  \\
\bottomrule
\end{tabular}
}
\end{table}

\begin{table}[t]
\centering
\caption{Ablation study on the loss weight $\alpha$.}
\label{tab:ablation_alpha}
\resizebox{\linewidth}{!}{
\begin{tabular}{c|ccc}
\toprule
$\boldsymbol{\alpha}$ &\textbf{Efficiency Acc. (\%)} &  \textbf{Toxic Acc. (\%)} & \textbf{MAE} \\
\midrule
0.00 &55.23\%  &90.41\%  &1.44  \\
0.05 &65.44\%  &90.30\%  &1.31  \\
\cellcolor[HTML]{DAE8FC}\textbf{0.10} &\cellcolor[HTML]{DAE8FC}\textbf{70.40\%}  &\cellcolor[HTML]{DAE8FC}\textbf{90.30\%} &\cellcolor[HTML]{DAE8FC}\textbf{1.24}  \\
0.15 &70.50\%  &88.63\%  &1.24   \\
0.20 &70.40\%  &86.32\%  &1.24   \\
\bottomrule
\end{tabular}
}
\end{table}

We conduct ablation studies to analyze two key design choices in \agentname{}:
(1) the timing of human-in-the-loop intervention during multi-agent verification, and
(2) the weighting factor $\alpha$ in the conditional multi-task loss.
All ablation experiments follow the setting in Section~\ref{sec:exp_setting}.

\paragraph{Human-in-the-loop intervention timing.}
As shown in Table~\ref{tab:ablation_loop}, the timing of human feedback critically affects accuracy, efficiency, and practical usability.
Introducing human feedback too early results in excessive manual review and limits the effectiveness of autonomous multi-agent reasoning.
Conversely, introducing it too late leads to prolonged agent disagreement, during which one agent may unduly influence the other, undermining reliable arbitration.
Triggering human intervention only after three unsuccessful verification loops provides the best balance between autonomy and safety in this safety-critical setting.

\paragraph{Effect of loss weight $\alpha$.}
Table~\ref{tab:ablation_alpha} reports ablation results for the loss weighting factor $\alpha$.
When $\alpha$ is too small, optimization is dominated by toxicity prediction, limiting gains in efficiency accuracy.
As $\alpha$ increases, efficiency prediction improves accordingly.
We find that $\alpha=0.1$ yields the most favorable trade-off, substantially improving efficiency prediction without degrading toxicity performance.
Larger values of $\alpha$ offer no additional benefit and may slightly compromise toxicity accuracy, highlighting the importance of balanced conditional optimization.

\subsection{Wet Experimental Validation}
\label{sec:wet_validation}
\begin{figure*}[t]
  \includegraphics[width=\textwidth]{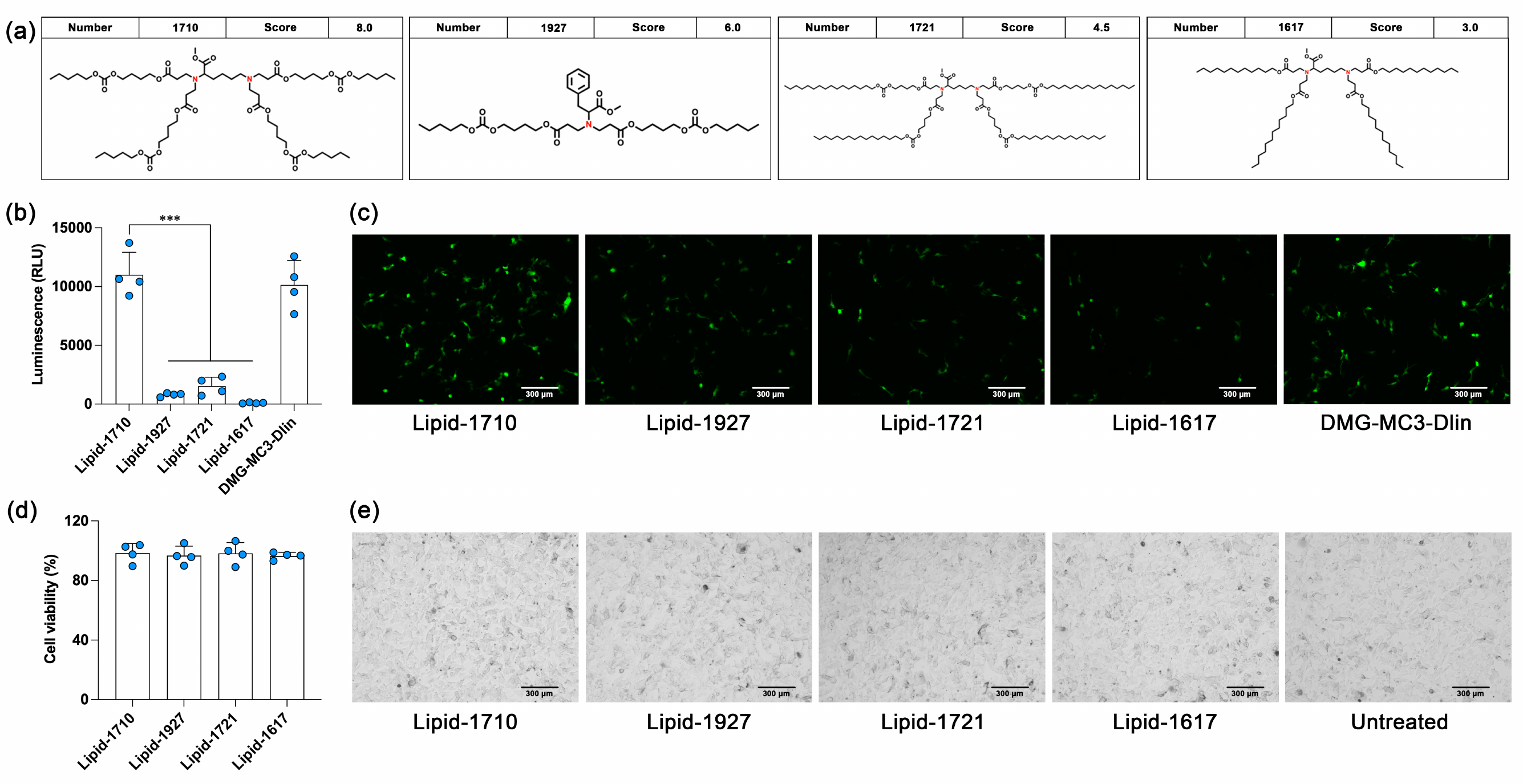}
  \vspace{-1.5em}
  \caption{In vitro evaluation of LNPs derived from predicted ionizable lipids.
  (a) The structures of lipids with different scores.
  (b) Luciferase (LUC) mRNA transfection efficiency of LNPs derived from the four predicted lipids in B16F10 cells.
  (c) Fluorescence imaging of B16F10 cells treated with EGFP mRNA-loaded LNPs derived from predicted lipids.
  (d) Cell viability of LNPs derived from predicted lipids evaluated by MTT assay.
  (e) Representative imaging of B16F10 cells treated with predicted lipid-derived LNPs.
  Scale bar: 300~$\mu$m.}
  \vspace{-1em}
  \label{fig:ann_exp_big}
\end{figure*}

To evaluate whether the predictions of \agentname{} translate into real biological performance, we conducted wet-lab experiments on lipid candidates selected via large-scale virtual screening.
We first construct a virtual lipid library containing 10{,}024 lipid molecules that are feasible to synthesize.
Using the best-performing configuration, \emph{TxGemma-27B with \agentname{}}, we perform \emph{in silico} screening over the entire library, jointly predicting toxicity and mRNA transfection efficiency for each lipid.

From the screening results, we selected \textbf{four} representative lipid candidates (Figure~\ref{fig:ann_exp_big}A) that are predicted to be non-toxic and span three distinct efficiency levels (high, medium, and low).
Importantly, this selection strategy is designed to validate the \emph{relative ordering} of model predictions across the efficiency spectrum, rather than to showcase only top-ranked candidates.
The detailed synthetic routes for all four selected lipid molecules are provided in Appendix~\ref{sec:appendix1}. LNPs are formulated using the ethanol injection method~\cite{eval_1} and evaluated \emph{in vitro} in B16F10 cells (Appendix~\ref{sec:appendix2}).

As shown in Figure~\ref{fig:ann_exp_big}B, Lipid-1710 exhibits the highest mRNA transfection efficiency among the four tested candidates, consistent with its predicted ranking.
The remaining three lipids demonstrate intermediate and lower efficiencies in accordance with their predicted relative ordering, supporting the robustness of the model’s ranking capability rather than isolated hit identification. Consistent trends were further confirmed by fluorescence imaging using EGFP mRNA (Figure~\ref{fig:ann_exp_big}C).

The model-generated explanations provide plausible structure--function hypotheses for the observed trends.
Compared with Lipid-1721, Lipid-1710 features an optimized hydrophobic tail length that may facilitate improved interactions with cellular membranes.
Relative to Lipid-1617, the presence of a biodegradable carbonate linkage in Lipid-1721 may promote more efficient intracellular mRNA release.
In comparison with Lipid-1927, the headgroup architecture and number of hydrophobic tails in Lipid-1710 appear more favorable for effective mRNA delivery.
Notably, under the same \emph{in vitro} experimental conditions, Lipid-1710 achieved mRNA transfection efficiency comparable to that of the commercially used lipid DMG-MC3-Dlin, serving as a benchmark.

MTT assays demonstrate negligible cytotoxicity for all four tested lipids (Figure~\ref{fig:ann_exp_big}D), and no significant differences in cell density or cellular morphology were observed between LNP-treated and untreated cells (Figure~\ref{fig:ann_exp_big}E).
Collectively, these results validate the reliability of \agentname{} in predicting relative mRNA transfection performance and highlight its potential to guide lipid design.

\subsection{Human Effort and Time Efficiency}
\label{sec:time_efficiency}

Beyond predictive accuracy, an important objective of \agentname{} is to reduce human effort required for lipid screening.
We compare our multi-agent framework with a conventional human-driven workflow, where experts synthesize large lipid libraries and experimentally evaluate candidates without automated filtering or agent-based verification.

In traditional pipelines, lipid synthesis is the dominant bottleneck.
In our wet-lab setting, synthesizing \textbf{four} lipids required approximately \textbf{96} hours of hands-on time, corresponding to about \textbf{24} hours per lipid.
Extrapolating to a virtual library of \textbf{10{,}024} candidates, exhaustive manual synthesis would require on the order of \textbf{240{,}000} hours, rendering experimental screening infeasible.

In contrast, \agentname{} completes \emph{in silico} screening of the full \textbf{10{,}024}-lipid library within approximately \textbf{23} hours and identifies the top \textbf{0.1\%} candidates (\textbf{10} lipids) for downstream validation.
Synthesizing these shortlisted lipids requires approximately \textbf{10} days, resulting in a total turnaround time of about \textbf{264} hours.
Overall, \agentname{} reduces the end-to-end time by approximately \textbf{99.89\%} while maintaining strict safety guarantees, as human supervision is invoked only for a small number of ambiguous cases.

\section{Conclusion}
We present LipoAgent, a safety-aware multi-agent LLM framework that jointly models toxicity and mRNA transfection efficiency for lipid discovery. Experiments and wet-lab validation show that LipoAgent can greatly improve prediction accuracy and reliability while ensuring safety-critical decision making.

\section{Limitations}
Despite its effectiveness, LipoAgent has several limitations.
First, the training data remain limited in scale and are aggregated from heterogeneous experimental sources, which may introduce residual noise despite manual normalization.
Second, mRNA transfection efficiency is modeled using discretized scores, which facilitates ranking but may obscure fine-grained quantitative differences between lipid candidates.
Third, toxicity assessment relies on available molecular toxicity datasets and in vitro assays, and does not fully capture complex in vivo safety profiles.
Fourth, LipoAgent is designed to predict the mRNA transfection efficiency of given lipid candidates rather than to directly generate novel, high-efficiency lipid molecules in an end-to-end manner.
Finally, although wet-lab experiments validate the predictive ordering of selected lipids, experimental evaluation is conducted on a limited number of candidates.
Future work will focus on expanding standardized datasets, improving continuous efficiency modeling, incorporating richer toxicity annotations, extending the framework toward generative lipid design, and scaling experimental validation.

\textbf{Potential Risks.} Although our framework is intended to support safer lipid candidate screening by prioritizing toxicity awareness, its predictions may still be inaccurate or miscalibrated outside the training distribution. As a result, model outputs should not be treated as definitive evidence of biological safety or experimental efficacy. The system is designed only as a decision-support tool for virtual screening and must be followed by expert review and wet-lab validation before any real-world use.

\clearpage

\bibliography{custom}

@article{hu2022lora,
  title={Lora: Low-rank adaptation of large language models.},
  author={Hu, Edward J and Shen, Yelong and Wallis, Phillip and Allen-Zhu, Zeyuan and Li, Yuanzhi and Wang, Shean and Wang, Lu and Chen, Weizhu and others},
  journal={ICLR},
  volume={1},
  number={2},
  pages={3},
  year={2022}
}

@misc{survey1,
      title={Large Language Models in Drug Discovery and Development: From Disease Mechanisms to Clinical Trials}, 
      author={Yizhen Zheng and Huan Yee Koh and Maddie Yang and Li Li and Lauren T. May and Geoffrey I. Webb and Shirui Pan and George Church},
      year={2024},
      eprint={2409.04481},
      archivePrefix={arXiv},
      primaryClass={q-bio.QM},
      url={https://arxiv.org/abs/2409.04481}, 
}

@article{survey2,
   title={DeepPurpose: a deep learning library for drug–target interaction prediction},
   volume={36},
   ISSN={1367-4811},
   url={http://dx.doi.org/10.1093/bioinformatics/btaa1005},
   DOI={10.1093/bioinformatics/btaa1005},
   number={22–23},
   journal={Bioinformatics},
   publisher={Oxford University Press (OUP)},
   author={Huang, Kexin and Fu, Tianfan and Glass, Lucas M and Zitnik, Marinka and Xiao, Cao and Sun, Jimeng},
   editor={Wren, Jonathan},
   year={2020},
   month=dec, pages={5545–5547} }

@misc{ReACT,
      title={ReAct: Synergizing Reasoning and Acting in Language Models}, 
      author={Shunyu Yao and Jeffrey Zhao and Dian Yu and Nan Du and Izhak Shafran and Karthik Narasimhan and Yuan Cao},
      year={2023},
      eprint={2210.03629},
      archivePrefix={arXiv},
      primaryClass={cs.CL},
      url={https://arxiv.org/abs/2210.03629}, 
}

@misc{Researchagent,
      title={ResearchAgent: Iterative Research Idea Generation over Scientific Literature with Large Language Models}, 
      author={Jinheon Baek and Sujay Kumar Jauhar and Silviu Cucerzan and Sung Ju Hwang},
      year={2025},
      eprint={2404.07738},
      archivePrefix={arXiv},
      primaryClass={cs.CL},
      url={https://arxiv.org/abs/2404.07738}, 
}

@misc{chemcrow,
      title={ChemCrow: Augmenting large-language models with chemistry tools}, 
      author={Andres M Bran and Sam Cox and Oliver Schilter and Carlo Baldassari and Andrew D White and Philippe Schwaller},
      year={2023},
      eprint={2304.05376},
      archivePrefix={arXiv},
      primaryClass={physics.chem-ph},
      url={https://arxiv.org/abs/2304.05376}, 
}

@misc{DrugAgent,
      title={DrugAgent: Automating AI-aided Drug Discovery Programming through LLM Multi-Agent Collaboration}, 
      author={Sizhe Liu and Yizhou Lu and Siyu Chen and Xiyang Hu and Jieyu Zhao and Yingzhou Lu and Yue Zhao},
      year={2025},
      eprint={2411.15692},
      archivePrefix={arXiv},
      primaryClass={cs.LG},
      url={https://arxiv.org/abs/2411.15692}, 
}

@article{safetysurvey,
  author       = {Ramos, Miguel and Li, Chen and Patel, Arjun and Zhang, Wei and Torres, Lucia},
  title        = {A review of large language models and autonomous agents in chemistry},
  journal      = {Chemical Science},
  year         = {2025},
  volume       = {16},
  number       = {4},
  pages        = {1234--1256},
  publisher    = {Royal Society of Chemistry},
  doi          = {10.1039/D4SC03921A},
  url          = {https://pubs.rsc.org/en/content/articlehtml/2025/sc/d4sc03921a},
}

@misc{qwen,
      title={Qwen3 Technical Report}, 
      author={An Yang and Anfeng Li and Baosong Yang and Beichen Zhang and Binyuan Hui and Bo Zheng and Bowen Yu and Chang Gao and Chengen Huang and Chenxu Lv and Chujie Zheng and Dayiheng Liu and Fan Zhou and Fei Huang and Feng Hu and Hao Ge and Haoran Wei and Huan Lin and Jialong Tang and Jian Yang and Jianhong Tu and Jianwei Zhang and Jianxin Yang and Jiaxi Yang and Jing Zhou and Jingren Zhou and Junyang Lin and Kai Dang and Keqin Bao and Kexin Yang and Le Yu and Lianghao Deng and Mei Li and Mingfeng Xue and Mingze Li and Pei Zhang and Peng Wang and Qin Zhu and Rui Men and Ruize Gao and Shixuan Liu and Shuang Luo and Tianhao Li and Tianyi Tang and Wenbiao Yin and Xingzhang Ren and Xinyu Wang and Xinyu Zhang and Xuancheng Ren and Yang Fan and Yang Su and Yichang Zhang and Yinger Zhang and Yu Wan and Yuqiong Liu and Zekun Wang and Zeyu Cui and Zhenru Zhang and Zhipeng Zhou and Zihan Qiu},
      year={2025},
      eprint={2505.09388},
      archivePrefix={arXiv},
      primaryClass={cs.CL},
      url={https://arxiv.org/abs/2505.09388}, 
}

@misc{CAMEL,
      title={CAMEL: Communicative Agents for "Mind" Exploration of Large Language Model Society}, 
      author={Guohao Li and Hasan Abed Al Kader Hammoud and Hani Itani and Dmitrii Khizbullin and Bernard Ghanem},
      year={2023},
      eprint={2303.17760},
      archivePrefix={arXiv},
      primaryClass={cs.AI},
      url={https://arxiv.org/abs/2303.17760}, 
}

@misc{Voyager,
      title={Voyager: An Open-Ended Embodied Agent with Large Language Models}, 
      author={Guanzhi Wang and Yuqi Xie and Yunfan Jiang and Ajay Mandlekar and Chaowei Xiao and Yuke Zhu and Linxi Fan and Anima Anandkumar},
      year={2023},
      eprint={2305.16291},
      archivePrefix={arXiv},
      primaryClass={cs.AI},
      url={https://arxiv.org/abs/2305.16291}, 
}

@misc{Reflexion,
      title={Reflexion: Language Agents with Verbal Reinforcement Learning}, 
      author={Noah Shinn and Federico Cassano and Edward Berman and Ashwin Gopinath and Karthik Narasimhan and Shunyu Yao},
      year={2023},
      eprint={2303.11366},
      archivePrefix={arXiv},
      primaryClass={cs.AI},
      url={https://arxiv.org/abs/2303.11366}, 
}

@misc{AGLE,
      title={AGILE platform: a deep learning powered approach to accelerate LNP development for mRNA delivery}, 
      author={Y. Xu and S. Ma and H. Cui and others},
      year={2024},
      eprint={2303.11366},
      archivePrefix={Nat Commun 15, 6305 (2024). },
      primaryClass={bio},
      url={https://doi.org/10.1038/s41467-024-50619-z}, 
}

@misc{LANTERN,
      title={LANTERN: A Machine Learning Framework for Lipid Nanoparticle Transfection Efficiency Prediction}, 
      author={Asal Mehradfar and Mohammad Shahab Sepehri and Jose Miguel Hernandez-Lobato and Glen S. Kwon and Mahdi Soltanolkotabi and Salman Avestimehr and Morteza Rasoulianboroujeni},
      year={2025},
      eprint={2507.03209},
      archivePrefix={arXiv},
      primaryClass={q-bio.QM},
      url={https://arxiv.org/abs/2507.03209}, 
}

@misc{KnowMol,
      title={KnowMol: Advancing Molecular Large Language Models with Multi-Level Chemical Knowledge}, 
      author={Zaifei Yang and Hong Chang and Ruibing Hou and Shiguang Shan and Xilin Chen},
      year={2025},
      eprint={2510.19484},
      archivePrefix={arXiv},
      primaryClass={q-bio.BM},
      url={https://arxiv.org/abs/2510.19484}, 
}

@misc{SCE,
      title={Scalable and Cost-Efficient de Novo Template-Based Molecular Generation}, 
      author={Piotr Gaiński and Oussama Boussif and Andrei Rekesh and Dmytro Shevchuk and Ali Parviz and Mike Tyers and Robert A. Batey and Michał Koziarski},
      year={2025},
      eprint={2506.19865},
      archivePrefix={arXiv},
      primaryClass={q-bio.BM},
      url={https://arxiv.org/abs/2506.19865}, 
}

@misc{DrugPilot,
      title={DrugPilot: LLM-based Parameterized Reasoning Agent for Drug Discovery}, 
      author={Kun Li and Zhennan Wu and Shoupeng Wang and Jia Wu and Shirui Pan and Wenbin Hu},
      year={2025},
      eprint={2505.13940},
      archivePrefix={arXiv},
      primaryClass={cs.AI},
      url={https://arxiv.org/abs/2505.13940}, 
}

@misc{ChemLLM,
      title={ChemLLM: A Chemical Large Language Model}, 
      author={Di Zhang and Wei Liu and Qian Tan and Jingdan Chen and Hang Yan and Yuliang Yan and Jiatong Li and Weiran Huang and Xiangyu Yue and Wanli Ouyang and Dongzhan Zhou and Shufei Zhang and Mao Su and Han-Sen Zhong and Yuqiang Li},
      year={2024},
      eprint={2402.06852},
      archivePrefix={arXiv},
      primaryClass={cs.AI},
      url={https://arxiv.org/abs/2402.06852}, 
}

@misc{Txgemma,
      title={TxGemma: Efficient and Agentic LLMs for Therapeutics}, 
      author={Eric Wang and Samuel Schmidgall and Paul F. Jaeger and Fan Zhang and Rory Pilgrim and Yossi Matias and Joelle Barral and David Fleet and Shekoofeh Azizi},
      year={2025},
      eprint={2504.06196},
      archivePrefix={arXiv},
      primaryClass={cs.AI},
      url={https://arxiv.org/abs/2504.06196}, 
}

@misc{Llama,
      title={The Llama 3 Herd of Models}, 
      author={Aaron Grattafiori and Abhimanyu Dubey and Abhinav Jauhri and Abhinav Pandey and Abhishek Kadian and Ahmad Al-Dahle and Aiesha Letman and Akhil Mathur and Alan Schelten and Alex Vaughan and Amy Yang and Angela Fan and Anirudh Goyal and Anthony Hartshorn and Aobo Yang and Archi Mitra and Archie Sravankumar and Artem Korenev and Arthur Hinsvark and Arun Rao and Aston Zhang and Aurelien Rodriguez and Austen Gregerson and Ava Spataru and Baptiste Roziere and Bethany Biron and Binh Tang and Bobbie Chern and Charlotte Caucheteux and Chaya Nayak and Chloe Bi and Chris Marra and Chris McConnell and Christian Keller and Christophe Touret and Chunyang Wu and Corinne Wong and Cristian Canton Ferrer and Cyrus Nikolaidis and Damien Allonsius and Daniel Song and Danielle Pintz and Danny Livshits and Danny Wyatt and David Esiobu and Dhruv Choudhary and Dhruv Mahajan and Diego Garcia-Olano and Diego Perino and Dieuwke Hupkes and Egor Lakomkin and Ehab AlBadawy and Elina Lobanova and Emily Dinan and Eric Michael Smith and Filip Radenovic and Francisco Guzmán and Frank Zhang and Gabriel Synnaeve and Gabrielle Lee and Georgia Lewis Anderson and Govind Thattai and Graeme Nail and Gregoire Mialon and Guan Pang and Guillem Cucurell and Hailey Nguyen and Hannah Korevaar and Hu Xu and Hugo Touvron and Iliyan Zarov and Imanol Arrieta Ibarra and Isabel Kloumann and Ishan Misra and Ivan Evtimov and Jack Zhang and Jade Copet and Jaewon Lee and Jan Geffert and Jana Vranes and Jason Park and Jay Mahadeokar and Jeet Shah and Jelmer van der Linde and Jennifer Billock and Jenny Hong and Jenya Lee and Jeremy Fu and Jianfeng Chi and Jianyu Huang and Jiawen Liu and Jie Wang and Jiecao Yu and Joanna Bitton and Joe Spisak and Jongsoo Park and Joseph Rocca and Joshua Johnstun and Joshua Saxe and Junteng Jia and Kalyan Vasuden Alwala and Karthik Prasad and Kartikeya Upasani and Kate Plawiak and Ke Li and Kenneth Heafield and Kevin Stone and Khalid El-Arini and Krithika Iyer and Kshitiz Malik and Kuenley Chiu and Kunal Bhalla and Kushal Lakhotia and Lauren Rantala-Yeary and Laurens van der Maaten and Lawrence Chen and Liang Tan and Liz Jenkins and Louis Martin and Lovish Madaan and Lubo Malo and Lukas Blecher and Lukas Landzaat and Luke de Oliveira and Madeline Muzzi and Mahesh Pasupuleti and Mannat Singh and Manohar Paluri and Marcin Kardas and Maria Tsimpoukelli and Mathew Oldham and Mathieu Rita and Maya Pavlova and Melanie Kambadur and Mike Lewis and Min Si and Mitesh Kumar Singh and Mona Hassan and Naman Goyal and Narjes Torabi and Nikolay Bashlykov and Nikolay Bogoychev and Niladri Chatterji and Ning Zhang and Olivier Duchenne and Onur Çelebi and Patrick Alrassy and Pengchuan Zhang and Pengwei Li and Petar Vasic and Peter Weng and Prajjwal Bhargava and Pratik Dubal and Praveen Krishnan and Punit Singh Koura and Puxin Xu and Qing He and Qingxiao Dong and Ragavan Srinivasan and Raj Ganapathy and Ramon Calderer and Ricardo Silveira Cabral and Robert Stojnic and Roberta Raileanu and Rohan Maheswari and Rohit Girdhar and Rohit Patel and Romain Sauvestre and Ronnie Polidoro and Roshan Sumbaly and Ross Taylor and Ruan Silva and Rui Hou and Rui Wang and Saghar Hosseini and Sahana Chennabasappa and Sanjay Singh and Sean Bell and Seohyun Sonia Kim and Sergey Edunov and Shaoliang Nie and Sharan Narang and Sharath Raparthy and Sheng Shen and Shengye Wan and Shruti Bhosale and Shun Zhang and Simon Vandenhende and Soumya Batra and Spencer Whitman and Sten Sootla and Stephane Collot and Suchin Gururangan and Sydney Borodinsky and Tamar Herman and Tara Fowler and Tarek Sheasha and Thomas Georgiou and Thomas Scialom and Tobias Speckbacher and Todor Mihaylov and Tong Xiao and Ujjwal Karn and Vedanuj Goswami and Vibhor Gupta and Vignesh Ramanathan and Viktor Kerkez and Vincent Gonguet and Virginie Do and Vish Vogeti and Vítor Albiero and Vladan Petrovic and Weiwei Chu and Wenhan Xiong and Wenyin Fu and Whitney Meers and Xavier Martinet and Xiaodong Wang and Xiaofang Wang and Xiaoqing Ellen Tan and Xide Xia and Xinfeng Xie and Xuchao Jia and Xuewei Wang and Yaelle Goldschlag and Yashesh Gaur and Yasmine Babaei and Yi Wen and Yiwen Song and Yuchen Zhang and Yue Li and Yuning Mao and Zacharie Delpierre Coudert and Zheng Yan and Zhengxing Chen and Zoe Papakipos and Aaditya Singh and Aayushi Srivastava and Abha Jain and Adam Kelsey and Adam Shajnfeld and Adithya Gangidi and Adolfo Victoria and Ahuva Goldstand and Ajay Menon and Ajay Sharma and Alex Boesenberg and Alexei Baevski and Allie Feinstein and Amanda Kallet and Amit Sangani and Amos Teo and Anam Yunus and Andrei Lupu and Andres Alvarado and Andrew Caples and Andrew Gu and Andrew Ho and Andrew Poulton and Andrew Ryan and Ankit Ramchandani and Annie Dong and Annie Franco and Anuj Goyal and Aparajita Saraf and Arkabandhu Chowdhury and Ashley Gabriel and Ashwin Bharambe and Assaf Eisenman and Azadeh Yazdan and Beau James and Ben Maurer and Benjamin Leonhardi and Bernie Huang and Beth Loyd and Beto De Paola and Bhargavi Paranjape and Bing Liu and Bo Wu and Boyu Ni and Braden Hancock and Bram Wasti and Brandon Spence and Brani Stojkovic and Brian Gamido and Britt Montalvo and Carl Parker and Carly Burton and Catalina Mejia and Ce Liu and Changhan Wang and Changkyu Kim and Chao Zhou and Chester Hu and Ching-Hsiang Chu and Chris Cai and Chris Tindal and Christoph Feichtenhofer and Cynthia Gao and Damon Civin and Dana Beaty and Daniel Kreymer and Daniel Li and David Adkins and David Xu and Davide Testuggine and Delia David and Devi Parikh and Diana Liskovich and Didem Foss and Dingkang Wang and Duc Le and Dustin Holland and Edward Dowling and Eissa Jamil and Elaine Montgomery and Eleonora Presani and Emily Hahn and Emily Wood and Eric-Tuan Le and Erik Brinkman and Esteban Arcaute and Evan Dunbar and Evan Smothers and Fei Sun and Felix Kreuk and Feng Tian and Filippos Kokkinos and Firat Ozgenel and Francesco Caggioni and Frank Kanayet and Frank Seide and Gabriela Medina Florez and Gabriella Schwarz and Gada Badeer and Georgia Swee and Gil Halpern and Grant Herman and Grigory Sizov and Guangyi and Zhang and Guna Lakshminarayanan and Hakan Inan and Hamid Shojanazeri and Han Zou and Hannah Wang and Hanwen Zha and Haroun Habeeb and Harrison Rudolph and Helen Suk and Henry Aspegren and Hunter Goldman and Hongyuan Zhan and Ibrahim Damlaj and Igor Molybog and Igor Tufanov and Ilias Leontiadis and Irina-Elena Veliche and Itai Gat and Jake Weissman and James Geboski and James Kohli and Janice Lam and Japhet Asher and Jean-Baptiste Gaya and Jeff Marcus and Jeff Tang and Jennifer Chan and Jenny Zhen and Jeremy Reizenstein and Jeremy Teboul and Jessica Zhong and Jian Jin and Jingyi Yang and Joe Cummings and Jon Carvill and Jon Shepard and Jonathan McPhie and Jonathan Torres and Josh Ginsburg and Junjie Wang and Kai Wu and Kam Hou U and Karan Saxena and Kartikay Khandelwal and Katayoun Zand and Kathy Matosich and Kaushik Veeraraghavan and Kelly Michelena and Keqian Li and Kiran Jagadeesh and Kun Huang and Kunal Chawla and Kyle Huang and Lailin Chen and Lakshya Garg and Lavender A and Leandro Silva and Lee Bell and Lei Zhang and Liangpeng Guo and Licheng Yu and Liron Moshkovich and Luca Wehrstedt and Madian Khabsa and Manav Avalani and Manish Bhatt and Martynas Mankus and Matan Hasson and Matthew Lennie and Matthias Reso and Maxim Groshev and Maxim Naumov and Maya Lathi and Meghan Keneally and Miao Liu and Michael L. Seltzer and Michal Valko and Michelle Restrepo and Mihir Patel and Mik Vyatskov and Mikayel Samvelyan and Mike Clark and Mike Macey and Mike Wang and Miquel Jubert Hermoso and Mo Metanat and Mohammad Rastegari and Munish Bansal and Nandhini Santhanam and Natascha Parks and Natasha White and Navyata Bawa and Nayan Singhal and Nick Egebo and Nicolas Usunier and Nikhil Mehta and Nikolay Pavlovich Laptev and Ning Dong and Norman Cheng and Oleg Chernoguz and Olivia Hart and Omkar Salpekar and Ozlem Kalinli and Parkin Kent and Parth Parekh and Paul Saab and Pavan Balaji and Pedro Rittner and Philip Bontrager and Pierre Roux and Piotr Dollar and Polina Zvyagina and Prashant Ratanchandani and Pritish Yuvraj and Qian Liang and Rachad Alao and Rachel Rodriguez and Rafi Ayub and Raghotham Murthy and Raghu Nayani and Rahul Mitra and Rangaprabhu Parthasarathy and Raymond Li and Rebekkah Hogan and Robin Battey and Rocky Wang and Russ Howes and Ruty Rinott and Sachin Mehta and Sachin Siby and Sai Jayesh Bondu and Samyak Datta and Sara Chugh and Sara Hunt and Sargun Dhillon and Sasha Sidorov and Satadru Pan and Saurabh Mahajan and Saurabh Verma and Seiji Yamamoto and Sharadh Ramaswamy and Shaun Lindsay and Shaun Lindsay and Sheng Feng and Shenghao Lin and Shengxin Cindy Zha and Shishir Patil and Shiva Shankar and Shuqiang Zhang and Shuqiang Zhang and Sinong Wang and Sneha Agarwal and Soji Sajuyigbe and Soumith Chintala and Stephanie Max and Stephen Chen and Steve Kehoe and Steve Satterfield and Sudarshan Govindaprasad and Sumit Gupta and Summer Deng and Sungmin Cho and Sunny Virk and Suraj Subramanian and Sy Choudhury and Sydney Goldman and Tal Remez and Tamar Glaser and Tamara Best and Thilo Koehler and Thomas Robinson and Tianhe Li and Tianjun Zhang and Tim Matthews and Timothy Chou and Tzook Shaked and Varun Vontimitta and Victoria Ajayi and Victoria Montanez and Vijai Mohan and Vinay Satish Kumar and Vishal Mangla and Vlad Ionescu and Vlad Poenaru and Vlad Tiberiu Mihailescu and Vladimir Ivanov and Wei Li and Wenchen Wang and Wenwen Jiang and Wes Bouaziz and Will Constable and Xiaocheng Tang and Xiaojian Wu and Xiaolan Wang and Xilun Wu and Xinbo Gao and Yaniv Kleinman and Yanjun Chen and Ye Hu and Ye Jia and Ye Qi and Yenda Li and Yilin Zhang and Ying Zhang and Yossi Adi and Youngjin Nam and Yu and Wang and Yu Zhao and Yuchen Hao and Yundi Qian and Yunlu Li and Yuzi He and Zach Rait and Zachary DeVito and Zef Rosnbrick and Zhaoduo Wen and Zhenyu Yang and Zhiwei Zhao and Zhiyu Ma},
      year={2024},
      eprint={2407.21783},
      archivePrefix={arXiv},
      primaryClass={cs.AI},
      url={https://arxiv.org/abs/2407.21783}, 
}

@article{dataset1,
  title   = {Artificial intelligence-driven rational design of ionizable lipids for mRNA delivery},
  author  = {Wang, Wei and Chen, Kai and Jiang, Tao and others},
  journal = {Nature Communications},
  volume  = {15},
  pages   = {10804},
  year    = {2024},
  doi     = {10.1038/s41467-024-55072-6},
  url     = {https://doi.org/10.1038/s41467-024-55072-6}
}

@article{dataset2,
  author       = {Lu, Jiang},
  title        = {The toxicity data of compounds},
  year         = {2024},
  publisher    = {figshare},
  doi          = {10.6084/m9.figshare.27195339.v5},
  url          = {https://doi.org/10.6084/m9.figshare.27195339.v5}
}

@article{ann_intro1,
  title     = {Non-viral vectors for RNA delivery},
  author    = {Yan, Yu and Liu, Xinyu and Lu, An and Wang, Xinyu and Jiang, Lixin and Wang, Jianchun},
  journal   = {Journal of Controlled Release},
  volume    = {342},
  pages     = {241--279},
  year      = {2022},
  month     = {February},
  doi       = {10.1016/j.jconrel.2022.01.008},
  pmid      = {35016918},
  pmcid     = {PMC8743282},
  issn      = {0168-3659},
  publisher = {Elsevier}
}

@article{an_intro2,
  title     = {Protection from previous natural infection compared with mRNA vaccination against SARS-CoV-2 infection and severe COVID-19 in Qatar: a retrospective cohort study},
  author    = {Chemaitelly, Hiam and Ayoub, Houssein H. and AlMukdad, Sarah and Coyle, Patrick and Tang, Patrick and Yassine, Hadi M. and Al-Khatib, Hiam A. and Smatti, Muna K. and Hasan, Mohammad R. and Al-Kanaani, Zaina and Al-Kuwari, Eman and Jeremijenko, Andrew and Kaleeckal, Aisha H. and Latif, Anvar N. and Shaik, Ranya M. and Abdul-Rahim, Hiam F. and Nasrallah, Ghada K. and Al-Kuwari, Maryam G. and Butt, Adeel A. and Al-Romaihi, Hanan E. and Al-Thani, Mohamed H. and Al-Khal, Abdul-Latif and Bertollini, Roberto and Abu-Raddad, Laith J.},
  journal   = {The Lancet Microbe},
  volume    = {3},
  number    = {12},
  pages     = {e944--e955},
  year      = {2022},
  month     = {December},
  doi       = {10.1016/S2666-5247(22)00287-7},
  pmid      = {36375482},
  pmcid     = {PMC9651957},
  publisher = {Elsevier}
}

@article{an_intro3,
  title     = {Lipid Nanoparticle (LNP) Enables mRNA Delivery for Cancer Therapy},
  author    = {Zong, Yu and Lin, Yifan and Wei, Tingting and Cheng, Qiang},
  journal   = {Advanced Materials},
  volume    = {35},
  number    = {51},
  pages     = {e2303261},
  year      = {2023},
  month     = {December},
  doi       = {10.1002/adma.202303261},
  pmid      = {37196221},
  publisher = {Wiley}
}

@article{an_intro5,
  title     = {Accelerating ionizable lipid discovery for mRNA delivery using machine learning and combinatorial chemistry},
  author    = {Li, Bo and Raji, Ibrahim O. and Gordon, Alexander G. R. and Sun, Lu and Raimondo, Thomas M. and Oladimeji, Fisayo A. and Jiang, A. Y. and Varley, Andrew and Langer, Robert S. and Anderson, Daniel G.},
  journal   = {Nature Materials},
  volume    = {23},
  number    = {7},
  pages     = {1002--1008},
  year      = {2024},
  month     = {July},
  doi       = {10.1038/s41563-024-01867-3},
  publisher = {Nature Publishing Group}
}

@article{ann_add,
  title   = {Double Braking Effects of Nanomedicine on Mitochondrial Permeability Transition Pore for Treating Idiopathic Pulmonary Fibrosis},
  author  = {Lu, A. and Xu, Z. and Zhao, Z. and Yan, Y. and Jiang, L. and Geng, J. and Jin, H. and Wang, X. and Liu, X. and Zhu, Y. and Shi, Y. and Liu, L. and Dai, H. and Wang, J. C.},
  journal = {Advanced Science},
  year    = {2024},
  volume  = {11},
  number  = {47},
  pages   = {e2405406},
  doi     = {10.1002/advs.202405406},
  pmid    = {39475000},
  pmcid   = {PMC11653616}
}

@article{bg_1,
  title   = {Rational design of lipid nanoparticles: overcoming physiological barriers for selective intracellular mRNA delivery},
  author  = {Zhao, Y. and Wang, Z. M. and Song, D. and Chen, M. and Xu, Q.},
  journal = {Current Opinion in Chemical Biology},
  year    = {2024},
  volume  = {81},
  pages   = {102499},
  doi     = {10.1016/j.cbpa.2024.102499},
  pmid    = {38996568},
  pmcid   = {PMC11323194}
}

@article{eval_1,
  title   = {Lipid nanoparticles: Composition, formulation, and application},
  author  = {Xu, S. and Hu, Z. and Song, F. and Xu, Y. and Han, X.},
  journal = {Molecular Therapy -- Methods \& Clinical Development},
  year    = {2025},
  volume  = {33},
  number  = {2},
  pages   = {101463},
  doi     = {10.1016/j.omtm.2025.101463},
  pmid    = {40927763},
  pmcid   = {PMC12415982}
}

@article{An_1,
  author  = {Zong, Y. and Lin, Y. and Wei, T. and Cheng, Q.},
  title   = {Lipid Nanoparticle (LNP) Enables mRNA Delivery for Cancer Therapy},
  journal = {Advanced Materials},
  year    = {2023},
  volume  = {35},
  number  = {51},
  pages   = {e2303261},
  doi     = {10.1002/adma.202303261},
  pmid    = {37196221}
}

@article{An_2,
  author  = {Zhang, Z. and Yao, S. and Wang, Y. and Luo, K. and Amiji, M. and Anderson, K. C.},
  title   = {Cancer vaccines: Discovery, development, and challenges for clinical translation},
  journal = {Biomaterials},
  year    = {2026},
  month   = feb,
  volume  = {325},
  pages   = {123615},
  doi     = {10.1016/j.biomaterials.2025.123615},
  pmid    = {40795707}
}

@article{An_3,
  author  = {Hou, Xiaohu and Zaks, Tal and Langer, Robert and Dong, Yizhou},
  title   = {Lipid nanoparticles for mRNA delivery},
  journal = {Nature Reviews Materials},
  year    = {2021},
  volume  = {6},
  pages   = {1078--1094},
  doi     = {10.1038/s41578-021-00358-0}
}

@article{An_4,
  author  = {Eygeris, Y. and Gupta, M. and Kim, J. and Sahay, G.},
  title   = {Chemistry of Lipid Nanoparticles for RNA Delivery},
  journal = {Accounts of Chemical Research},
  year    = {2022},
  month   = jan,
  volume  = {55},
  number  = {1},
  pages   = {2--12},
  doi     = {10.1021/acs.accounts.1c00544},
  pmid    = {34850635}
}

@article{An_5,
  author  = {Hald Albertsen, Christian and Kulkarni, Jayesh A. and Witzigmann, Dominik and Lind, M. and Petersson, Karsten and Simonsen, Jens B.},
  title   = {The role of lipid components in lipid nanoparticles for vaccines and gene therapy},
  journal = {Advanced Drug Delivery Reviews},
  year    = {2022},
  month   = sep,
  volume  = {188},
  pages   = {114416},
  doi     = {10.1016/j.addr.2022.114416},
  pmid    = {35787388},
  pmcid   = {PMC9250827}
}

@article{An_6,
  author  = {Su, K. and Qiu, J. and Xu, T. and Liu, S.},
  title   = {Artificial intelligence-guided design of lipid nanoparticles for mRNA delivery},
  journal = {Acta Pharmaceutica Sinica B},
  year    = {2026},
  month   = feb,
  volume  = {16},
  number  = {2},
  pages   = {709--727},
  doi     = {10.1016/j.apsb.2025.11.029},
  pmid    = {41685167},
  pmcid   = {PMC12891898}
}

@article{lnp_mrna1,
  author  = {Sun, D. and Lu, Z. R.},
  title   = {Structure and Function of Cationic and Ionizable Lipids for Nucleic Acid Delivery},
  journal = {Pharmaceutical Research},
  year    = {2023},
  month   = jan,
  volume  = {40},
  number  = {1},
  pages   = {27--46},
  doi     = {10.1007/s11095-022-03460-2},
  pmid    = {36600047},
  pmcid   = {PMC9812548}
}

@article{lnp_mrna2,
  author  = {Xu, Y. and Golubovic, A. and Xu, S. and Pan, A. and Li, B.},
  title   = {Rational design and combinatorial chemistry of ionizable lipids for RNA delivery},
  journal = {Journal of Materials Chemistry B},
  year    = {2023},
  month   = jul,
  volume  = {11},
  number  = {28},
  pages   = {6527--6539},
  doi     = {10.1039/d3tb00649b},
  pmid    = {37345430}
}

@article{LNP_AI1,
  author  = {Su, Kexin and Qiu, Junjie and Xu, Tengfei and Liu, Shuai},
  title   = {Artificial intelligence-guided design of lipid nanoparticles for mRNA delivery},
  journal = {Acta Pharmaceutica Sinica B},
  year    = {2026},
  volume  = {16},
  number  = {2},
  pages   = {709--727},
  doi     = {10.1016/j.apsb.2025.11.029}
}

@article{LNP_AI2,
  author  = {Xu, Y. and Golubovic, A. and Xu, S. and Pan, A. and Li, B.},
  title   = {Rational design and combinatorial chemistry of ionizable lipids for RNA delivery},
  journal = {Journal of Materials Chemistry B},
  year    = {2023},
  volume  = {11},
  number  = {28},
  pages   = {6527--6539}
}

@article{AI_Safety1,
  author  = {Hanna, A. R. and Issadore, D. and Mitchell, M. J.},
  title   = {High-throughput platforms for machine learning-guided lipid nanoparticle design},
  journal = {Nature Reviews Materials},
  year    = {2025}
}

@article{AI_Safety2,
  author  = {Wang, J. and Ding, Y. and Chong, K. and Cui, M. and Cao, Z. and Tang, C. and Tian, Z. and Hu, Y. and Zhao, Y. and Jiang, S.},
  title   = {Recent Advances in Lipid Nanoparticles and Their Safety Concerns for mRNA Delivery},
  journal = {Vaccines},
  year    = {2024},
  month   = oct,
  volume  = {12},
  number  = {10},
  pages   = {1148},
  doi     = {10.3390/vaccines12101148},
  pmid    = {39460315},
  pmcid   = {PMC11510967}
}

@inproceedings{AI4S1,
  author    = {Gang Liu and Michael Sun and Wojciech Matusik and Meng Jiang and Jie Chen},
  title     = {Multimodal Large Language Models for Inverse Molecular Design with Retrosynthetic Planning},
  booktitle = {International Conference on Learning Representations (ICLR)},
  year      = {2025}
}

@inproceedings{AI4S2,
  author    = {Tung Nguyen and Aditya Grover},
  title     = {LICO: Large Language Models for In-Context Molecular Optimization},
  booktitle = {International Conference on Learning Representations (ICLR)},
  year      = {2025}
}

@inproceedings{AI4S3,
  author    = {Weitong Zhang and Xiaoyun Wang and Weili Nie and Joe Eaton and Brad Rees and Quanquan Gu},
  title     = {MoleculeGPT: Instruction Following Large Language Models for Molecular Property Prediction},
  booktitle = {NeurIPS 2023 Workshop on New Frontiers of AI for Drug Discovery and Development},
  year      = {2023}
}

\clearpage

\appendix
\section{Synthetic Routes of Selected Lipids}
\label{sec:appendix1}
\begin{figure}[h]
  \includegraphics[width=1.0\columnwidth]{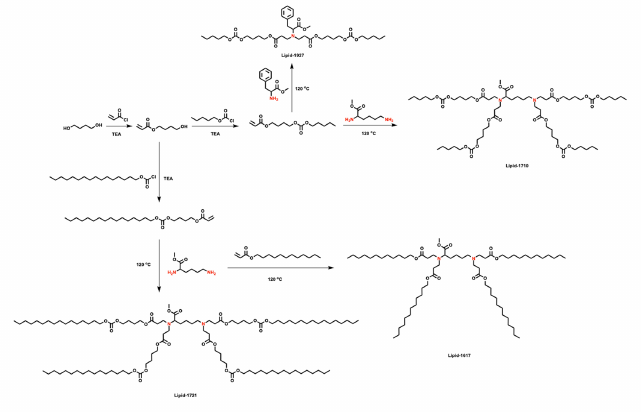}
  \vspace{-1.5em}
    \caption{The synthetic routes of four selected lipids.}
\vspace{-1em}
  \label{fig:ann_exp1}
\end{figure}
\paragraph{Lipid synthesis.}
The chemical structures and synthetic routes of the selected lipids are illustrated in Figure~\ref{fig:ann_exp1}.
Specifically, Lipid-1710 is synthesized as follows.
A mixture of butyleneglycol, acryloyl chloride, and triethylamine is dissolved in tetrahydrofuran (THF) and stirred at room temperature for 30 minutes.
After solvent removal under reduced pressure, the crude product is purified by column chromatography to obtain intermediate aT1.
Subsequently, aT1 is reacted with pentyl chloroformate in dichloromethane (DCM) at room temperature for 30 minutes, followed by solvent removal and column chromatography to yield intermediate T1.
Finally, T1 is reacted with H-Lys-OMe in methanol at 120~$^\circ$C for 4 hours, and the product is purified by column chromatography to obtain Lipid-1710.

Lipid-1927 is synthesized by reacting intermediate T1 with H-Phe-OMe in methanol at 120~$^\circ$C for 4 hours, followed by solvent removal and column chromatography purification.
For Lipid-1721, intermediate aT1 is first reacted with hexadecyl carbonochloridate in DCM at room temperature for 30 minutes to generate intermediate T2.
T2 is then reacted with H-Lys-OMe in methanol at 120~$^\circ$C for 4 hours, followed by purification to obtain Lipid-1721.
Lipid-1617 is synthesized by directly reacting dodecyl acrylate with H-Lys-OMe in methanol at 120~$^\circ$C for 4 hours, followed by purification.

\section{Experimental Protocols for LNP Preparation and Evaluation}
\label{sec:appendix2}
\begin{figure}[h]
  \includegraphics[width=1.0\columnwidth]{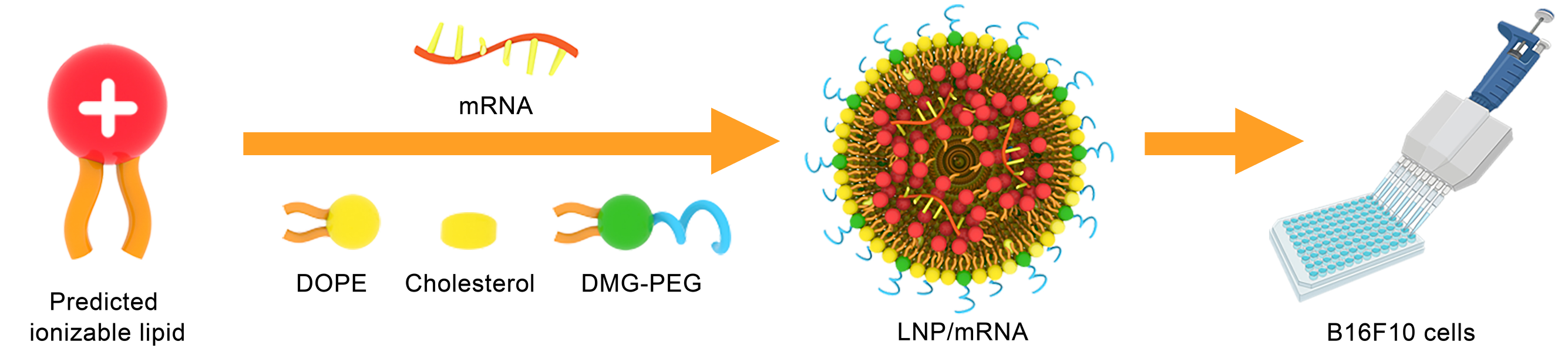}
  \vspace{-1.5em}
    \caption{The schematic illustration of LNP preparation and in vitro cellular evaluation procedures.}
\vspace{-1em}
  \label{fig:ann_exp2}
\end{figure}

As shown in Figure~\ref{fig:ann_exp2}, the synthesized lipid is used to prepare LNPs loading mRNA, and then the prepared LNPs are evaluated using B16F10 cells. luciferase (LUC) reporter assay and EGFP reporter assay are used to detect the in  vitro mRNA transfection. MTT assay is used to detect the cytotoxicity of LNPs.

\paragraph{Preparation of lipid nanoparticles (LNPs).}
For LNP formulation, mRNA is first dissolved in citrate buffer (pH~=~4) to obtain solution~A.
Ionizable lipid, DOPE, cholesterol (Chol), and DMG-PEG2000 are dissolved in ethanol to obtain solution~B.
Solution~B is then mixed with solution~A at a volume ratio of B/A~=~1/3.
The resulting mixture is dialyzed overnight against RNase-free PBS using a dialysis membrane with a molecular weight cutoff of 10{,}000--12{,}000~Da to obtain stable LNP formulations.

\paragraph{Transfection efficiency evaluation.}
mRNA transfection efficiency is evaluated using B16F10 cells.
Cells are seeded into 96-well plates at a density of 20{,}000 cells per well and incubated with 100~$\mu$L of DMEM medium containing LUC mRNA-loaded LNPs for 24 hours.
Luciferase expression levels are quantified using the Firefly Luciferase Reporter Gene Assay Kit, with the final LUC mRNA concentration fixed at 100~ng per well.
For imaging-based evaluation, B16F10 cells are treated with enhanced green fluorescent protein (EGFP) mRNA-loaded LNPs for 24 hours and subsequently observed under a fluorescence microscope.

\paragraph{MTT assay.}
B16F10 cells (20000 cells per well) were seeded into 96-well plates and incubated with 100 uL DMEM medium containing mRNA-loaded LNPs for 24 h. Then CyQUANTTM MTT Cell Proliferation Assay Kit was used to detect the cytotoxicity of LNPs.

\end{document}